\documentclass[10pt,twocolumn,letterpaper]{article}

\usepackage{wacv}              

\usepackage{wrapfig}
\usepackage[table]{}
\usepackage{tabularx}
\usepackage{amsmath, dsfont} 

\usepackage{xcolor, colortbl}
\definecolor{Gray}{gray}{0.9}
\definecolor{LightCyan}{rgb}{0.88,1,1}
\definecolor{golden}{rgb}{1,0.84, 0}

\usepackage{graphicx}
\usepackage{amsmath}
\usepackage{amssymb}
\usepackage{booktabs}

\usepackage{nicefrac,xfrac}

\newcommand{\methodname}{RISE}

\usepackage[pagebackref,breaklinks,colorlinks]{hyperref}

\usepackage[capitalize]{cleveref}
\crefname{section}{Sec.}{Secs.}
\Crefname{section}{Section}{Sections}
\Crefname{table}{Table}{Tables}
\crefname{table}{Tab.}{Tabs.}

\def\wacvPaperID{2524} 
\def\confName{WACV}
\def\confYear{2025}

\begin{document}

\title{Robot Instance Segmentation with Few
Annotations for Grasping}


\makeatletter
\renewcommand{\@fnsymbol}[1]{\ifcase#1\or $\dagger$\or 1\or 2\or 3\or 5\or 6\or 7\or 8\or 9\fi}

\author{Moshe Kimhi\thanks{Equal contribution}~~\thanks{Technion – Israel Institute of Technology}~~\thanks{Bosch Center for AI}   \and
David Vainshtein\footnotemark[1]~~\footnotemark[3] \and
Chaim Baskin \thanks{Ben-Gurion University of the Negev} 
\and Dotan Di Castro \footnotemark[3]}

\maketitle

\begin{abstract}
The ability of robots to manipulate objects relies heavily on their aptitude for visual perception. In domains characterized by cluttered scenes and high object variability such as traffic, navigation and object grasping, most methods call for vast labeled datasets, laboriously hand-annotated, with the aim of training capable models. Once deployed, the challenge of generalizing to unfamiliar objects implies that the model must evolve alongside its domain. To address this, we propose a novel framework that combines Semi-Supervised Learning (SSL) with Learning Through Interaction (LTI), allowing a model to learn by observing scene alterations and leverage visual consistency despite temporal gaps without requiring curated data of interaction sequences. As a result, our approach exploits partially annotated data through self-supervision and incorporates temporal context using pseudo-sequences generated from unlabeled still images.
We validate our method on two common benchmarks, ARMBench mix-object-tote and OCID, where it achieves state-of-the-art performance. Notably, on ARMBench, we attain an $\text{AP}_{50}$ of $86.37$, almost a $20\%$ improvement over existing work, and obtain remarkable results in scenarios with extremely low annotation, achieving an $\text{AP}_{50}$ score of $84.89$ with just $1 \%$ of annotated data compared to previous state of the art of $82$ which targeted the fully annotated dataset.
\end{abstract}

\section{Introduction}
\label{sec:intro}


\noindent 
Acquiring accurate instance segmentation masks requires training a model on vast amounts of data with high-quality pixel-level annotations. While collecting raw sensory data (images) is relatively easy, annotating object instance masks down to individual pixels becomes prohibitively expensive when scaling up perception tasks. As a result, models trained on limited annotated data inevitably face challenges when deployed in the real world due to domain variation and evolving environments. This problem is central in robotics, where robots rely on spatial perception extracted from sensory inputs.

\begin{figure}[t!]
  \centering
  \setlength{\belowcaptionskip}{-10pt}
   \includegraphics[width=0.7\linewidth]{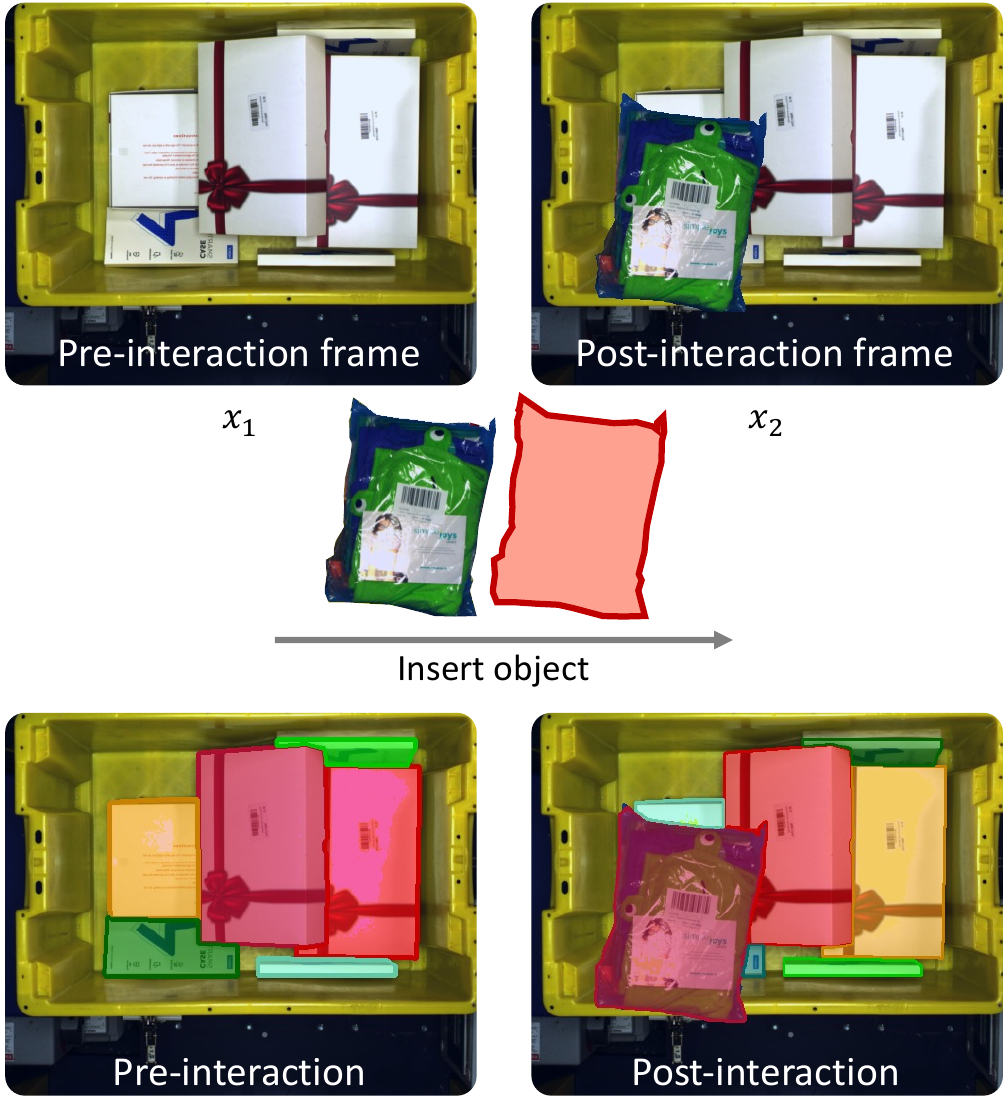}
\caption{Pseudo-sequence generation from a single unlabeled image. The input is weakly augmented to produce the ``before'' and ``after'' images $x_1$ and $x_2$. To emulate scene interaction, objects are drawn from the object memory bank, transformed, and inserted into the ``after'' frame. The segmentation model's task is to simultaneously associate objects that persist between frames (subject to occlusion), maintain consistency of object instance embedding, and correctly predict the ground-truth mask of the added objects.}
\label{fig:rise_overview}
\end{figure}

To use large amounts of unlabeled data, \textbf{Semi-Supervised Learning (SSL)} assumes that only a portion of the data is labeled: either a subset of observed scenes or some objects within each scene. The model then uses its own predictions as pseudo-labels to extract learning signals from the remaining unlabeled data \cite{sohn2020fixmatch,xie2020uda,zhang2021flexmatch,wang2023freematch,unimatch}. Therefore, a model attempting to learn from its own noisy labels early in training may stagnate rather than generalize.



Looking beyond spatial cues of still images, video sequences contain temporal information that a model can exploit to enforce consistency across frames and improve generalization. Recent advancements focusing on \textbf{Learning Through Interaction (LTI)} highlight the significance of providing the model with temporal perception. LTI enables the model to peer into the underlying dynamics of its domain by observing actions and their consequences \cite{Kroemer2019ARO,Garg2021SemanticsFR,tang2023graspgpt}. The leading approaches entail observing a scene that undergoes various changes, such as objects being placed or extracted. By constructing the data in the form of ``before'' and ``after'' sequences \cite{seqformer,IDOL,ying2023ctvis,stow,Liu2023SelfSupervisedIS}, localized changes in illumination, deformation, and articulation of objects allow the model to refine its interpretation of the environment. The leading LTI techniques either prescribe multi-stage training that pre-trains on specialized datasets \cite{Liu2023SelfSupervisedIS} or restrict input observations to strictly gradual changes at small time intervals \cite{chen2022unsupervised}.
Interestingly, leading methods for Video Image Segmentation regularly resolve long image sequences depicting instances popping in and out of view \cite{seqformer, IDOL, ying2023ctvis, maskfreevis}. These incorporate a reassociation loss to overcome changes of occlusion, instance pose and appearance. However, learning from videos relies either on significant investment in manual annotation of every object in every video frame or on video frames occurring at sufficiently small time intervals.
Our main insight is that although each paradigm compensates for the weakness of the other (SSL lessens the annotation effort while LTI leverages temporal information), naively combining the two amplifies their drawbacks---reinforcing noisy labels across entire sequences.
In this work, we propose a solution in the form of a novel framework that incorporates the learning paradigms of SSL and LTI to enhance performance in the few-annotations scenario, in which only a tiny fraction of the dataset is annotated (and the rest is unlabeled).
Our method simultaneously addresses the challenges of LTI and SSL. We eliminate the need for specialized datasets required for LTI by using pseudo-sequences generated from still images to mimic scene interaction. We also overcome the main obstacles to SSL by preventing noisy self-predictions from obscuring the learning signal through coupling prediction heads, thus stabilizing predictions early in training.
Our framework is model-agnostic, complementing existing (and future) segmentation models with temporal perception through end-to-end training.
Additionally, we propose an automated pseudo-label criteria that discards low-quality predictions. 

The resulting framework can be considered the first to employ self-supervised learning through interaction, achieving better performance than each paradigm individually. 
We set a new state of the art on the ARMBench \cite{Mitash2023ARMBenchAO} benchmark and OCID \cite{OCID} (RGB only). Notably, our method trained on $1\%$ of annotated data surpasses the performance of the well-established Deformable DETR \cite{zhu2020deformable} architecture, even when trained on $10\times$ additional annotated data (improving  $+16.86$ AP using Swin-L Transformer as feature extractor).

\section{Background and Related Work}
\label{sec:related}

\noindent Of the various approaches to instance segmentation, we are interested in those that excel without full supervision \cite{Kroemer2019ARO,Garg2021SemanticsFR}. This section provides an overview of relevant works on partial supervision and learning from sequences.

\textbf{Partial supervision} methods use the few annotated examples available (if any) and maintain consistent predictions for similar objects in the scene \cite{Zoph2020RethinkingPA}. In recent years, most efforts focused on contrastive learning that extracts embedding from object instances and aims to bring same-class embedding closer while pushing other classes further apart \cite{chen2020simclrv2, sohn2020fixmatch}. That said, progress in object classification and detection does not readily carry over to image segmentation, where the effectiveness of self-supervision lags behind full-supervision in challenging domains of cluttered objects with many occlusions \cite{ziegler2022self}. Unsurprisingly, these domains are also more complicated for humans to annotate. 

\textbf{Scene modulation} is a concept that aims to extract additional learning signal by familiarizing the model with objects that undergo gradual alterations within a scene \cite{wen2021catgrasp,xie2021unseen,stow,Wen2022TransGraspGP, dwibedi2017cut} where objects are viewed in many configurations, as well as different clutter and lighting conditions. This offers a substantial advantage in detecting and identifying objects that may deform or exhibit variations, thereby enhancing the robustness of the segmentation. Note, however, that assembling large dedicated datasets of objects is resource-intensive and challenging to apply effectively to new scenes featuring previously unseen objects. A recent work \cite{xie2021unseen} achieved significant improvement by incorporating simulated data before transferring to real world scenes \cite{Horvth2022Sim2RealGP,stow}. The main drawback of using synthetic data is the high cost of creating photorealistic rendering that accurately captures the physical properties of every object in the scene. Often times this results in idiosyncrasies that are picked up by the model and become a source of error when encountering real world data.

\textbf{Frame sequences} offer additional information along the time dimension. As with scene modulation, the model learns to recognize and identify related instances throughout a series of images \cite{chen2022unsupervised}. Recent advancements in video instance segmentation (VIS) methods, exemplified by SeqFormer \cite{seqformer} and IDOL \cite{IDOL}, leverage sequential consistency of instances for online object segmentation and tracking. They employ contrastive loss to ensure that instance representations are distinguishable from other instances in the same frame and over previous frames. In CTVIS \cite{ying2023ctvis} the model also taps into future frames.

\textbf{Learning through interaction} pushes the notion of sequences even further by specializing in image sequences that depict predefined and controlled scene manipulation. Consecutive frames in these meticulously assembled datasets exhibit large temporal gaps, unlike video data, and changes are usually confined to localized actions on few object instances \cite{Liu2023SelfSupervisedIS}. This locality constraint persists through frame sequences, allowing LTI approaches to infer which instances have actually changed and which are merely affected by variations in lighting, occlusion and deformation, as a result of the action performed. The model quickly learns to segment an object that is added or removed, using a few hundred labeled image pairs. Since assembling such specialized datasets requires significant effort, the next stage in training artificially inserts cropped instances from high-confidence mask predictions into unlabeled still-images to emulate interactions. In STOW \cite{stow}, the model is additionally trained on synthesized virtual scenes and then evaluated on real-world data.

The above advancements present an interesting question: 
In real-world applications where the model inevitably encounters a changing domain, is it possible to continuously learn (post deployment) without supervision by leveraging the temporal information of video sequences using the causal awareness of LTI?\\
Importantly, can this be achieved without investing in a proprietary dataset or reliance on a specific instance segmentation model?
\section{\methodname{}}
\label{sec:method}
We introduce a novel framework called \textbf{R}obot \textbf{I}nstance \textbf{Se}gmentation for Few-Annotation Grasping (\textbf{\methodname}) that unifies learning from temporal signals (through interactions) and spatial signals (through self-supervision). \methodname~is trained end-to-end on still images, and enables self-supervision to learn from temporal consistency when scene objects are moved, added or removed. Because of this, \methodname~does not require a meticulously compiled dataset of before and after image pairs of scene interaction, nor does it require a large dataset of labeled instances --- thus it is more readily capable of handling domain variations that commonly occur in the real world. 

\begin{figure*}[ht!]
  \centering
  \setlength{\belowcaptionskip}{-10pt}
   \includegraphics[width=\linewidth]{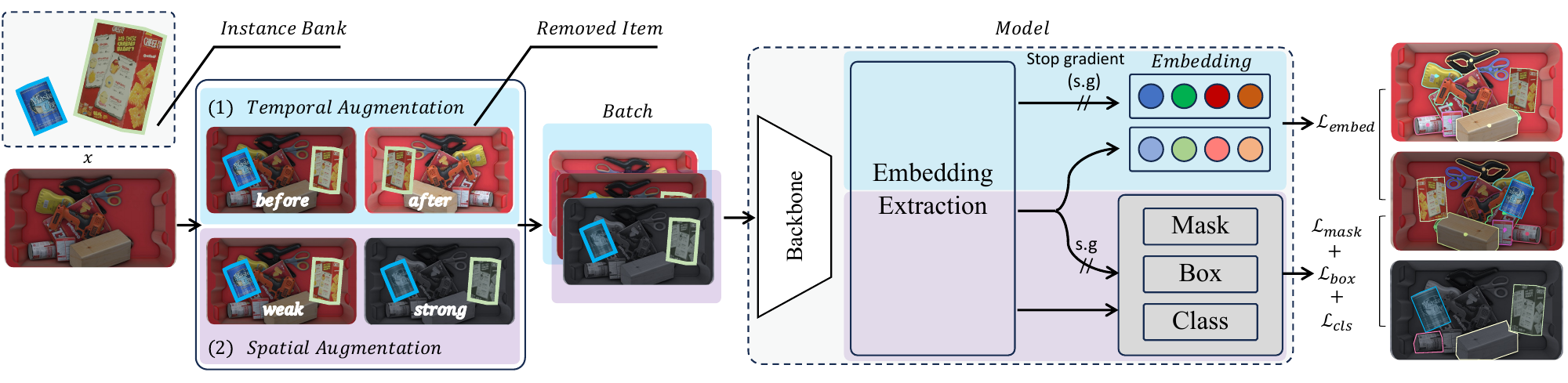}
   \caption{\methodname~framework (from left to right). Given an unlabeled image $x$ and a bank of known instances, 
   we perform (1) temporal  (\textcolor{cyan}{blue}) and (2) spatial (\textcolor{violet}{purple}) augmentations. \textcolor{cyan}{Temporal augmentation} adds weak augmentation and inserts $K$ instances from the bank to create $x_1$ (the ``before'' frame). Another round of weak augmentations, combined with adding/moving/removing a subset of the $K$ added instances, produces $x_2$ (the ``after'' frame).
   \textcolor{violet}{Spatial augmentations} adds strong augmentations to create $x_3$. The three images are batched and fed into the model, where the backbone extracts features that are then encoded into instance embedding. The instance embedding from $x_1$ and $x_2$ are used to compute $\mathcal{L}_{embed}$ \cref{eq:loss_embed}. The embedding from $x_1$ serve as pseudo-labels against the embedding from $x_3$ in the self-supervised loss $\mathcal{L}_\mathrm{u}$ \cref{eq:unsupervised_loss}.}
   \label{fig:rise_architecture}
\end{figure*}

The framework accommodates both supervised and semi-supervised data, comprising an instance segmentation model (\cref{sec:instance_segmentation}) that is enhanced to extract learning signals from both instance-association and consistency losses (\cref{subsec:association_loss}).
\subsection{Instance Segmentation}
\label{sec:instance_segmentation}
Object instance segmentation begins with the input image $x$ which first undergoes feature extraction by an extractor backbone. The features are then fed into an instance level embedding encoder that outputs $300$ tokens that serve the decoder, which emits instance embedding $z_i$ into the predictions heads for class, bounding box and mask of object instance $i$.
In this work we evaluate various established backbones: Resnet50, Resnet101 \cite{resnet}, and Swin-L \cite{liu2021Swin} as options for the feature extractor.
As embedding decoder we chose Deformable DETR \cite{zhu2020deformable} as a strong spatial decoder for its ability to learn object queries as features.
The prediction heads for class labels and box coordinates are feed-forward networks (FFNs), whereas the mask prediction head is a Feature-Pyramid network (FPN) \cite{FPN} that uses multi-scale features from the decoder's last layers, followed by an FFN whose output mask is scaled up to match the original image size.

When the input is also accompanied by labels $\mathbf{y}$, the supervised component of the loss constitutes a class label loss $\mathcal{L}_{cls}$; $\mathcal{L}_{box}$ that combines $L_1$ loss and generalized Intersection over Union (gIoU) loss \cite{rezatofighi2019generalized}; $\mathcal{L}_{mask}$ as the sum of the Dice loss \cite{milletari2016v} and Focal loss \cite{lin2017focal}:
\begin{equation}
\label{eq:supervised_loss} 
 \mathcal{L}_{\mathrm{s}} = \mathcal{L}_{cls}+\lambda_1 \mathcal {L}_{box}+ \lambda_2 \mathcal{L}_{mask},
\end{equation}
where $\lambda_1$ and $\lambda_2$ are the loss coefficients.

Recent advancements in object detection incorporated optimal transport (OT) to address the optimal assignment between predictions and ground truth, as proposed by Ge et al. \cite{Ge2021OTAOT} and YOLOX \cite{yolox2021}. Therefore, we compute the pairwise cost between predictions and ground truth instances, determining the optimal assignment for the top $k$ predictions associated with each instance.

While operating on labeled data, object instances that are successfully segmented by the model are stored in a memory bank \cite{ghiasi2020simple}. This object bank will be used in the self-supervised phase, during which labeled instances are randomly selected and augmented to emulate scene interaction.
\subsection{Learning Through Interaction} 
\label{subsec:association_loss}
Observing interactions shares commonality with Object Tracking, which goes beyond traditional object detection. It leverages discriminative representation of instances across frames and of different instances belonging to the same class. The resulting representation is more robust to occlusion and identity switches, as demonstrated in \cite{1641019, 1211433}.

Given an unlabeled input frame $x$ containing an unknown number of object instances, we introduce a new augmentation strategy to create a pair of pre- and post-interaction frames. The first, pre-interaction frame $x_1=\psi_1(x)$ is an augmentation $\psi_1$ of $x$ where we stochastically insert $K$ objects from a bank of known instances. Each of the $K$ objects is also individually augmented (e.g., scale, position, rotation, flip, color). The second, post-interaction frame $x_2=\psi_2(x_1)$ is an augmentation $\psi_2$ of $x_1$ where we also remove several of the objects added to $x_1$ or insert a few more objects from the instance bank (with augmentations). Note that both $x_1$ and $x_2$ are spatially augmented with rotation, crop, and scale to convey a sense of motion to the observer (inspired by SeqFormer \cite{seqformer}).
\smallskip\\
\textbf{Augmentation Strategy}
\label{par:augmentation_strategy}
Inserting new objects into a dense scene may lead to significant occlusions and even conceal the objects we intend to learn. Therefore, we devise a strategy that randomizes labeled objects from the instance bank and distributes them preferentially around the periphery of the frame:
\begin{align}
\label{eq:beta}
   (u,v) = Beta(\alpha,\beta) \cdot [w, h]
\end{align}
where $(u,v)$ is the top-left corner where the object is inserted, drawn from distribution $Beta(\alpha, \beta) \in \mathbb{R}^2$, and $w$, $h$ are the feasible horizontal and vertical regions that ensure that the object is contained within the frame (see \cref{appdx:technical}). The choice of $Beta$ and its parameters reduces the likelihood of objects inserted near the center, where they might obstruct unlabeled objects.
Another safeguard prevents inserting an object if it would overlap with any of the previously inserted objects by more than $85\%$.
\smallskip\\
\textbf{Association Loss}
\label{par:Association_Loss}
The resulting frames $x_1$, $x_2$ contain a total of $N$ and $M$ object instances, respectively. Importantly, the objects' small projective and illumination transformations compel the model to learn robust representations that maintain consistency for occurrences of the same instance in a changing scene (illustrated in ~\cref{fig:rise_architecture}). 
Each embedding $i \in N$ extracted from the first frame $x_1$ is matched against every embedding $j \in M$ in the second frame $x_2$, forming the association score $f(i,j)$ between instance $i$ and instance $j$:
\begin{equation}
\label{eq:instance_score}
    f(i,j)=\frac{1}{2}\left[
    \frac{ \exp(z_j^T\cdot z_i)}{\sum\limits_{k=1}^M \exp(z_k^T\cdot z_i)}+
    \frac{\exp(z_j^T\cdot z_i)}{\sum\limits_{k=1}^N \exp(z_j^T\cdot z_k)}
    \right]
\end{equation}

We consider the embedding of $\hat{j} = \mathrm{arg\,max}\;f(i,j)$ as a positive example for the given instance $i$ if $f(i,\hat{j}) > 0.5$, otherwise it is considered a negative example.
We employ an embedding contrastive loss \cite{chen2020simple} to learn object representation from the observed interaction frames:
\begin{equation}
\label{eq:loss_embed}
    \mathcal{L}_{embed} = -\log \frac{\exp(z_i\cdot z_j^+)}
    {\exp(z_i\cdot z_j^+) +
    \sum_{z_i^{\!-}} \exp(z_i\cdot  z_j^{\!-})}
\end{equation}
where $z_i$ is the embedding of instance $i$ in the first frame, $z_j^+$ is the embedding of the instance $j$ in the second frame that ideally represents the same instance, and $z_j^{\!-}$ are the embedding of the remaining instances (negative views). The loss $\mathcal{L}_{embed}$ pulls same-instance embedding closer together while pushing apart representations of different instances.
\subsection{Self-Supervision}\label{sec:self_supervision}
To better leverage spatial information in unlabeled data, we employ the segmentation model (\cref{sec:instance_segmentation}) toward Semi-Supervised Learning (SSL). Inspired by \cite{sohn2020fixmatch}, we include a consistency regularization loss and extend it to accept unlabeled images alongside labeled objects inserted from the instance bank.

Recall that $x_1 = \psi_1(x)$ is a weak augmentation of the unlabeled input image $x$. In this context, we'll denote $x_w = x_1$. We apply another round of weak augmentations to $x_w$, followed by a strong augmentation $\phi$ to produce $x_s = x_3 = \phi(x_1)$. The strong augmentations comprise Color jitter, Planckian jitter \cite{zini2022planckian}, Gaussian blur, and gray-scale that are applied via RandAugment~\cite{Cubuk2019RandAugmentPD}.
We feed both $x_w$ and $x_s$ into the model. Class labels, bounding boxes, and segmentation masks for weakly augmented inputs $x_w$ are treated as pseudo-label targets (in the absence of ground truth) that are compared against the model's prediction on $x_s$. The unsupervised consistency regularization loss:
\begin{equation}
\label{eq:unsupervised_loss} 
 \mathcal{L}_{\mathrm{u}} = \hat{\mathcal{L}}_{cls}+\lambda_1 \hat{\mathcal{L}}_{box}+ \lambda_2 \hat{\mathcal{L}}_{mask}
\end{equation}
It is similar to the supervised loss $\mathcal{L}_{\mathrm{s}}$ (\cref{eq:supervised_loss}), with the distinction that pseudo-labels are used in place of ground-truth labels. Gradients are not computed during the forward pass of $x_w$ (as illustrated in \cref{fig:rise_architecture}) as it constitutes the ground truth.
We introduce the following refinements to stabilize the model during self-supervised training.
\smallskip\\
\textbf{Refined Consistency Learning}
\label{paragraph:consistency_learning}
It is common practice to filter out pseudo-labels with low prediction scores in order to reduce the model's exposure to errors during self-supervised training. The filters are often thresholds or quantiles that are either fixed, dynamic, or scheduled \cite{wang2022semi, Kimhi2023SemiSupervisedSS}. Thresholds, by nature, are more restrictive, discarding all predictions below their stated value. However, during the early stages of self-supervision, the model may emit most of its predictions slightly below the threshold, resulting in very few labels contributing towards learning. On the other hand, quantiles ignore the scores entirely and allow any prediction, provided that its score meets the rank requirement of the quantile. Because most models output a fixed number of predictions to accommodate crowded scenes (regularly exceeding $300$ predictions), a quantile may become too lenient and include low-score predictions of poor quality, potentially degrading the model's performance as training progresses. 

In \cref{sec:dynamic_threshold}, we demonstrate that early in training, setting the quantile too low lets in more predictions of low-quality signals, interfering with the model. Alternatively, setting the bar (too) high \cite{sohn2020fixmatch} risks missing out on meaningful supervision signals.

To reconcile the limitations of both thresholds and quantiles, we propose a cascade approach. First, a more relaxed class threshold $\gamma_t^{cls}$ removes instances whose class scores $\hat{c}_i \in \mathbf{\hat{c}}$ are deemed unusable, followed by a quantile selection $Q$ \cite{wang2022semi} of the leading predictions. The resulting class pseudo-labels $\hat{\mathbf{y}}$ is given by:
\begin{align}
\label{eq:quantile_threshold}
    \hat{\mathbf{y}} = Q(~\mathbf{\hat{c}} > 
         \gamma^{cls}_t~;~p_t).
\end{align}
The threshold $\gamma_t^{cls}$ discards instances with class scores $\hat{c}_i$ below it and tightens over time (training steps $t$).
Conversely, the quantile $Q(p_t)$ loosens over time with its probability  decays over subsequent training steps $t$, with $T$ denoting the total number of training steps. As a result, the quantile allows more predictions into the model as training progresses.
This strategy can mitigate incorrect model beliefs and reduce confirmation biases. We evaluate this strategy quantitatively and demonstrate its advantage over thresholds and quantiles in \cref{tab:abalation_threshold_cascade}, with additional details in \cref{appx:threshold_info}.
We recognize that exploring different quantile strategies may further improve self-supervision and set it aside for future work.
\smallskip\\
\textbf{Coupled Prediction Heads}
\label{paragrph:coupled_heads}
The standard approach to filtering pseudo-mask predictions employs a pixel-wise confidence threshold $\gamma_t^{mask}$ that is applied to each pixel $(u,v)$ of instance mask $\hat{m}_i$:
\begin{align}
\label{eqn:pseudomasks}
    \hat{m}^{u,v}_i =
    \begin{cases} 
        1 &\text{if } h^{\text{mask}}(z^w_i)  > \gamma^{\text{mask}}_t, \\
        0 &\text{otherwise}
    \end{cases}
\end{align}
where $h^{mask}$ is the mask head output for instance embedding $z_i^w$ obtained from the weakly augmented frame $x^w$.

Unlike masks, the prediction quality of bounding boxes is less correlated with high label scores. As such, recent SSL methods for object detection employ multiple passes to refine box predictions \cite{xu2021softteacher,Chen2022LabelMatching}.
Interestingly, we observe that the model learns to predict high quality masks well before it effectively predicts bounding boxes. Thus, we propose a coupling of the mask and box prediction heads so that during training, pseudo-boxes $\hat{b}_i$ are obtained by bounding their corresponding instance segmentation masks $\hat{m}_i$:
\begin{align}
\label{eqn:pseudo_box}
\hat{b}_i =  
\begin{bmatrix}
\displaystyle\min_u \hat{m}_i & 
\displaystyle\min_v \hat{m}_i & 
\displaystyle\max_u \hat{m}_i & 
\displaystyle\max_v \hat{m}_i
\end{bmatrix}.
\end{align}
We refer to this simple yet effective technique for pseudo-box assignment as Mask-to-Box (M2B) and demonstrate its advantage over the standard approach to predicting pseudo-boxes in  \cref{tab:abalation_components,tab:ablation_box_threshold}.
\smallskip\\
\textbf{Multi-Label Matching}
\label{paragrph:multi_label_matching}
A common practice in object segmentation is to apply non-maximum suppression (NMS) to eliminate redundant predictions. In our case, instance overlaps are common since objects are inserted at random as part of pseudo-sequence generation. Therefore, to make better use of the model's predictions during training, we introduce a new adaptation of Label-Matching (LM) \cite{Chen2022LabelMatching}, whereby we retain several overlapping predictions that coincide with the dominant class label (instead of discarding all but one). We call this method Multi-Label Matching (MLM) and conduct an ablation study to assess its contribution to self-supervision (\cref{tab:abalation_components,tab:ablation_box_threshold}), demonstrating its advantage.

\subsection{Unified Framework}
\label{sec:summrised_method}
The complete architecture of \methodname{} is presented in \cref{fig:rise_architecture}, comprising an LTI branch and an SSL branch that converge into a unified loss:
\begin{equation}
\label{eq:total_loss} 
 \mathcal{L}_{\text{total}} = \mathds{1}[\mathbf{y} \neq \varnothing] \mathcal{L}_{\mathrm{s}} + \lambda_3 \mathcal{L}_{\mathrm{embed}} + \mathds{1}[\mathbf{y} = \varnothing] \lambda_4 \mathcal{L}_{\mathrm{u}},
\end{equation}
where $\mathds{1}$ indicates that the supervised loss $\mathcal{L}_\mathrm{s}$ and unsupervised loss $\mathcal{L}_\mathrm{u}$ are used according to the availability of ground-truth labels $\mathbf{y}$, and $\mathcal{L}_{embed}$ denotes the weighted combinations of the association loss. Hyperparameter search for $\lambda_3$ and details on $\lambda_1, \lambda_2$ and $\lambda_4$ are provided in \cref{appdx:technical}.

\section{Experiments}
\label{sec:experiments}
We conduct a series of experiments to evaluate the performance of the proposed approach in the Robotic Item Grasping domain. This domain is of high relevance to automated distribution warehouses, where robotic arms pick and place items inside totes. The experiments target a range of labeled data ratios, meaning that we intentionally restrict the model's access to only a certain portion ($\%$) of the labeled samples, and treat the remaining samples as unlabeled.

\subsection{Setup}
\label{sec:setup}
\textbf{Datasets}
\label{paragrph:datasets}
Our main focus is the ARMBench \cite{Mitash2023ARMBenchAO} mix-tote benchmark comprising $44,\!234$ images, split into $30,\!992$ training images and $6,\!637$ and $6,\!605$ images for validation and testing, respectively. The images are not organized into sequences nor do they describe a localized action. Every object in the scene belongs to a single ``object'' category and is associated with a manually annotated instance mask.

\noindent The OCID \cite{OCID} dataset (containing  $2,\!390$ images and $31$ classes) for various rates of labeled-to-unlabeled data, and compare it to the current state of the art±\cite{msmformer}. 

\noindent We use the same \methodname{} configuration (e.g., Beta function, thresholds, etc.) for both datasets. The results in \cref{tab:ocid} illustrate that our method is readily applied to new datasets without requiring domain-specific configuration adjustments.
\smallskip\\
\textbf{Evaluation}
\label{par:eval}
We evaluate our method using the standard Average Precision (AP). We measure the overall AP across $10$ IoU thresholds $[0.50,\dots,0.95]$, as well as the IoU thresholded precision $\text{AP}_{50}$ and $\text{AP}_{75}$.
For OCID we use only the $\text{AP}_{50}$ to be consistent with prior art.

\noindent In terms of partitioning, we use $100\%$, $10\%$, $2\%$, $1\%$ and $0.5\%$  of the data as fully annotated, and the remaining as unlabeled for ARMBench, and $100\%$, $10\%$ and $5\%$ for OCID.
We compare \methodname{} with the officially reported performance from the ARMBench \cite{Mitash2023ARMBenchAO} and RoboLLM \cite{long2023robollm}, in which a model was trained on the entire training set. This baseline is the existing state-of-the-art on the ARMBench dataset. In addition, we compare RISE with Deformable DETR \cite{zong2022detrs} (denoted DeDETR).

\begin{table*}[h]
\centering
\caption{
ARMBench \cite{Mitash2023ARMBenchAO} mix-object-tote instance segmentation with subset of annotations. The first column denote the number of annotated samples, with the rest treated as unlabeled data for self-supervision. Second column details the method name and the next columns are $\text{AP}$ measures, with best performers marked in \textbf{bold}.  
}
\label{tab:classic}
\resizebox{0.7\linewidth}{!}{
\begin{tabular}{clccccccccc}
\toprule
 \rowcolor{Gray}$\%$&&\multicolumn{3}{c}{ResNet-50}&\multicolumn{3}{c}{ResNet-101}&\multicolumn{3}{c}{ViT}\\
  \rowcolor{Gray} Labeled  & Method  &$\text{AP}$&$\text{AP}_{50}$&$\text{AP}_{75}$ &$\text{AP}$&$\text{AP}_{50}$&$\text{AP}_{75}$ &$\text{AP}$&$\text{AP}_{50}$&$\text{AP}_{75}$\\
\midrule
\midrule
$0.5 \%$ 
& DeDETR\cite{zhu2020deformable} & 27.03 & 29.32 & 26.65 & 28.36 & 30.69 & 28.14 & 36.47 & 36.46 & 31.75 \\   
$(155)$ 

& M2F \cite{cheng2021mask2former} & - & - & - & - & - & - & 55.3 & 59.9  & 54.7  \\ 

& SAM \cite{kirillov2023segment} & - & - & - & - & - & - & 61.38 & 74.04 & 63.51 \\  
& 
\cellcolor{golden!20} \methodname{} & \cellcolor{golden!20} \textbf{66.15} & \cellcolor{golden!20} \textbf{78.80} &
\cellcolor{golden!20} \textbf{69.67} &
\cellcolor{golden!20} \textbf{71.40} &
\cellcolor{golden!20} \textbf{82.10} &
\cellcolor{golden!20} \textbf{72.30} &
\cellcolor{golden!20} \textbf{72.14} &
\cellcolor{golden!20} \textbf{83.25} &
\cellcolor{golden!20} \textbf{73.73} \\ 
\midrule

$1 \%$ 
& DeDETR \cite{zhu2020deformable} & 27.17 & 29.64 & 26.67 &30.38 & 34.52 & 29.73 & 39.46 & 39.44 & 33.51\\ 
$(309)$

& YOLACT  & - & - & - & 36.1 & 59.2 & 44.8 & - & - & - \\ 

& M2F \cite{cheng2021mask2former} & - & - & - & - & - & - & 58.6 & 64.7 & 58.6 \\

& SAM \cite{kirillov2023segment}  & - & - & - & - & - & - & 67.42 & 82.26 & 70.93 \\  
& 
\cellcolor{golden!20}\methodname{} &
\cellcolor{golden!20} \textbf{69.10} &
\cellcolor{golden!20} \textbf{82.10} &
\cellcolor{golden!20} \textbf{73.80} &
\cellcolor{golden!20} \textbf{73.00} &
\cellcolor{golden!20} \textbf{83.25} &
\cellcolor{golden!20} \textbf{73.94} & 
\cellcolor{golden!20} \textbf{73.72} & 
\cellcolor{golden!20} \textbf{84.89} & 
\cellcolor{golden!20} \textbf{74.89}  \\ 
\midrule

$ 2 \% $ 
& DeDETR\cite{zhu2020deformable} & 31.79 & 36.5 & 31.81 & 33.14 & 39.70 & 34.19 & 42.15 & 66.20  & 43.39 \\  

& M2F \cite{cheng2021mask2former} & - & - & - & - & - & - & 61.4 & 68.5 & 61.2 \\ 

$ (618) $ & 
\cellcolor{golden!20}  \methodname{} & 
\cellcolor{golden!20} \textbf{72.80} &
\cellcolor{golden!20} \textbf{82.90} & 
\cellcolor{golden!20} \textbf{74.40} & 
\cellcolor{golden!20} \textbf{73.66} & 
\cellcolor{golden!20} \textbf{83.44} & 
\cellcolor{golden!20} \textbf{75.89} & 
\cellcolor{golden!20} \textbf{73.92} & 
\cellcolor{golden!20} \textbf{84.00} & 
\cellcolor{golden!20} \textbf{76.25} \\  
\midrule

$10 \%$

& DeDETR \cite{zhu2020deformable}  &48.00 &57.44 &48.17 & 52.23 & 59.98  & 49.8 & 59.19 & 75.5  & 60.42 \\  
$(3,\!099)$ 
& YOLACT  & - & - & - & 47.40 & 68.20 & 52.70 & - & - & - \\ 

& M2F \cite{cheng2021mask2former} & - & - & - & - & - & - & 68.2 & 76.5 & 68.2 \\ 

& SAM \cite{kirillov2023segment} & - & - & - & - & - & - & 71.47 & 82.78 & 73.96 \\ 
& 
\cellcolor{golden!20} \methodname{} &
\cellcolor{golden!20} \textbf{73.39} & 
\cellcolor{golden!20} \textbf{83.48} &
\cellcolor{golden!20} \textbf{75.09} & 
\cellcolor{golden!20} \textbf{74.27} & 
\cellcolor{golden!20} \textbf{84.33} &
\cellcolor{golden!20} \textbf{75.54} &
\cellcolor{golden!20} \textbf{74.95} & 
\cellcolor{golden!20} \textbf{85.16} &
\cellcolor{golden!20} \textbf{76.26} \\
\midrule
$100 \%$ & ARMBench \cite{Mitash2023ARMBenchAO}  & - & 72.00 & 61.00 &-&-&-&-&-&- \\
$(30,\!992)$ 
& DeDETR \cite{zhu2020deformable}  & 52.11 &60.38&52.52 & 53.80 & 62.00  & 52.80 & 62.75 & 77.03  & 63.40 \\  

& M2F \cite{cheng2021mask2former} & - & - & - & - & - & - & 73.00 & 81.2 & 74.00 \\

& RoboLLM \cite{long2023robollm}  & - & - & - & - & - & - & - & 82.0 & 67.00 \\ 

&  
\cellcolor{golden!20} \methodname{} & \cellcolor{golden!20} \textbf{73.41} &
\cellcolor{golden!20} \textbf{83.53} & 
\cellcolor{golden!20} \textbf{75.15} & 
\cellcolor{golden!20} \textbf{74.47} &
\cellcolor{golden!20} \textbf{84.74} &
\cellcolor{golden!20} \textbf{75.93} &
\cellcolor{golden!20} \textbf{76.04} &
\cellcolor{golden!20} \textbf{86.37} &
\cellcolor{golden!20} \textbf{77.51}\\ 
\midrule
\bottomrule
\end{tabular}
}
\end{table*}

\subsection{Results}
\label{sec:results}
\begin{figure}[tb!]
\centering
\includegraphics[width=.5\textwidth]
{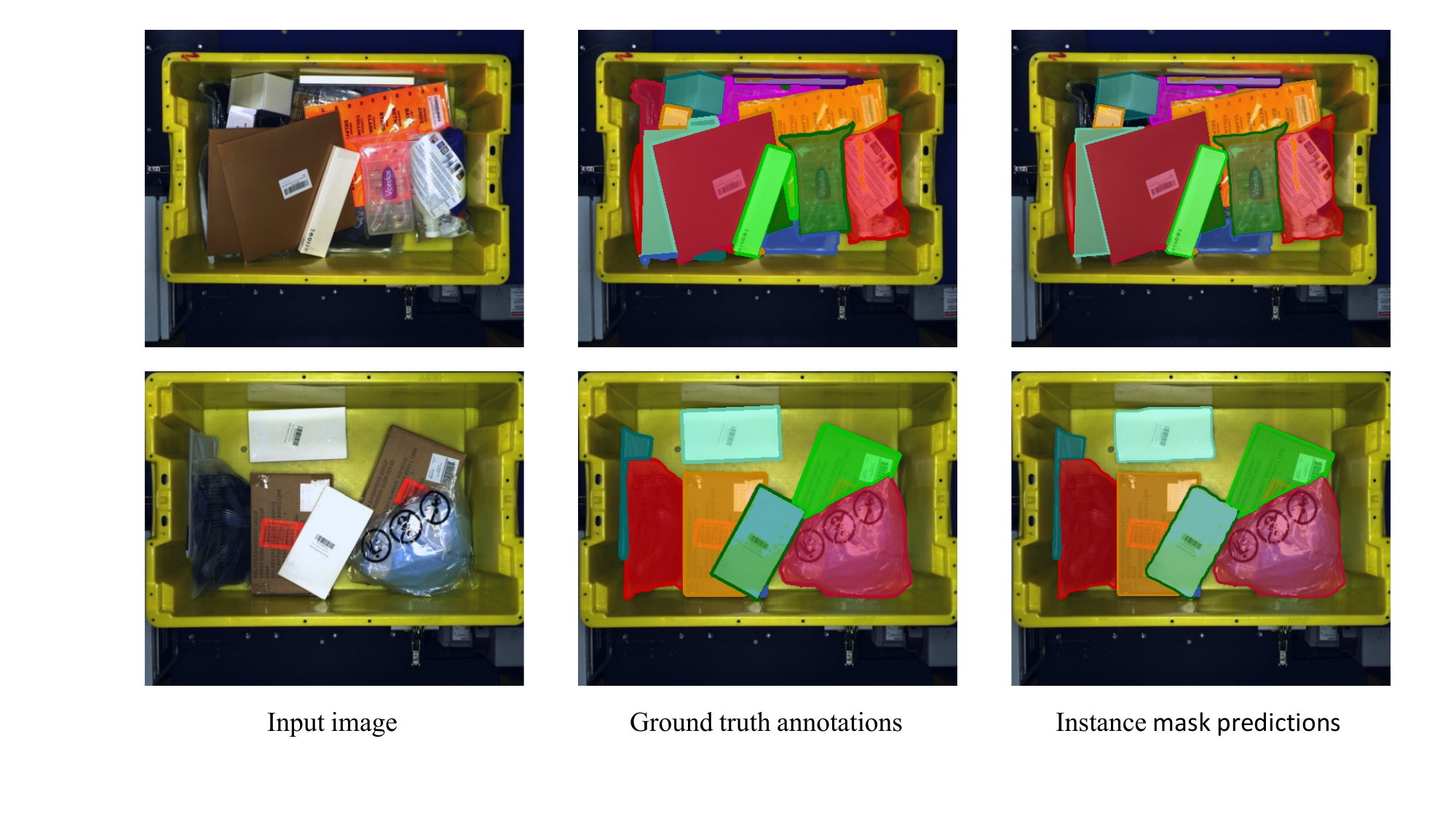}
\caption{Qualitative results of \methodname{} using ResNet-101 backbone, trained on $1\%$ of the labeled data ($99\%$ treated as unlabeled). Comparing the ground truth (center column) and the predicted masks (right-most column), we see that the majority of large items are accurately segmented, while some of the smaller or heavily occluded objects are occasionally missed. The segmentation masks are continuous, indicating high confidence for every object. Mask boundaries are the only regions with instance--background ambiguity (evident by mild noise at  object boundaries).}
\label{fig:qualitive_output}
\end{figure}

\cref{tab:classic} shows the results of \methodname{} on various data partitions of labeled/unlabeled ratios of the ARMBench, compared with Deformable DETR and SAM\cite{kirillov2023segany} (fine-tuned), as well as the results reported by the authors of ARMBench \cite{Mitash2023ARMBenchAO}. 
Both DeDETR and \methodname{} use Swin-L\cite{liu2021Swin} ($197$M parameters) as backbone, while SAM uses ViT-H\cite{li2022exploring} ($636$M parameters) and RoboLLM\cite{long2023robollm} uses Beit-3 base (87M parameters). Across all partitions, \methodname{} outperforms the other methods. \cref{fig:qualitive_output} illustrates high-quality masks predicted by \methodname{} trained on $1\%$ of the labeled, with $99\%$ of the remaining data treated as unlabeled. Most of the line-of-sight objects are accurately segmented and a few heavily occluded objects are missed. Importantly, \methodname{} trained on $1\%$ annotated data performs better than DeDETR and SAM trained on $10\%$ annotated samples ($10\times$ the amount of annotations for training/fine-tuning).
As an additional baseline we compare \methodname{} with the ``Segment Anything'' (SAM) \cite{kirillov2023segany}, for more details refer to \cref{paragraph:foundation_models}

\cref{tab:ocid} compares performance with partitions of labeled/unlabeled data ratios from OCID, showing the advantage of \methodname{}.

\begin{table}
\centering
\caption{
Evaluation on OCID\cite{OCID} (RGB only). $^+$ denotes Second stage networks. Bottom rows show the performance when only a portion ($\%$) of annotations are used. The proposed approach achieves significantly better results even with few annotations compared to prior art.}
\begin{tabular}{lc}
\toprule
Method &$\text{AP}_{50}$\\
\midrule
\midrule
UCN \cite{Xiang2020LearningRF} & 54.8 \\
UCN$^+$ \cite{Xiang2020LearningRF} & 59.1 \\
Mask2Former \cite{cheng2021mask2former} & 67.2 \\
MSMFormer\cite{Lu2022MeanSM} & 72.9 \\
MSMFormer$^+$ \cite{Lu2022MeanSM} & 73.9 \\
MRCNN \cite{He2017MaskR} & 77.6              \\
\rowcolor{golden!20}\methodname{}  & \textbf{78.2}   \\ 
\midrule
\rowcolor{golden!20}\methodname{} ($5\%$) & 75.1    \\ 
\rowcolor{golden!20}\methodname{} ($10\%$) & 77.3    \\ 
\midrule
\bottomrule
\end{tabular}
\label{tab:ocid}
\end{table}

\subsection{Ablation Study}
\label{sec:ablation}
We provide an ablation study of the various design choices made in implementing \methodname{}: impact of losses, pseudo-label threshold strategies and parameters. \cref{tab:abalation_components} details the contribution of the different elements within \methodname{} on the fully-supervised training set. The most substantial improvement is attributed to the Pseudo-Sequence (PS)  strategy outlined in \cref{subsec:association_loss}. The coupling of prediction heads in Mask-to-Box (M2B in \cref{eqn:pseudomasks}) refines the supervision signal for box predictions, further improving the performance. Combining it with Multi-Label Matching (MLM) and Optimal Transport (OT) yields the best performing version of \methodname{}.

\begin{table}
\centering
\caption{Ablation study on different components of \methodname{}. Pseudo-Sequences (PS), Optimal Transport (OT), Mask-to-Box coupling (M2B) and Multi-Label Matching (MLM). We evaluate the contribution of these component using $100\%$ of the annotated data, showing that the combined approach achieves the best results.}
\setlength{\tabcolsep}{6pt}
\label{tab:abalation_components}
\resizebox{0.8\linewidth}{!}{
\begin{tabular}{cccc|c}
\toprule
PS & OT & M2B & MLM & \text{AP} \\
\toprule
  &   &   &  &    53.80 \\ 
\midrule
\checkmark &   &   &   &  73.85 \\
\midrule
 & \checkmark  &   &   &  62.92 \\
\midrule
\checkmark & \checkmark  &   &  &    74.16 \\
\midrule
\checkmark & \checkmark  & \checkmark  &  &     74.35 \\
\midrule
\checkmark & \checkmark  & \checkmark  & \checkmark &   \textbf{74.47} \\
\bottomrule
\end{tabular}
}
\end{table}

\begin{table}
\centering
\caption{Ablation study of pseudo-label threshold strategy, comparing the score threshold $\gamma_t^{cls}$, the quantile $Q(p_t)$, and the proposed score filtering cascade (\cref{eq:quantile_threshold}) that applies a threshold followed by a quantile $\gamma_t^{cls} \rightarrow Q(p_t)$ (or reversed $ Q(p_t) \rightarrow \gamma_t^{cls}$). The best performance is achieved for the proposed cascade approach.}
\label{tab:abalation_threshold_cascade}
\resizebox{0.85\linewidth}{!}{
\begin{tabular}{l|ccc}
\toprule
 Threshold Strategy  & $\text{AP}$ & $\text{AP}_{50}$  & $\text{AP}_{75}$ \\
\midrule
Threshold only $\gamma_t^{cls}$ & 72.91 & 83.39 & 74.38 \\
Quantile only $Q(p_t)$ & 72.59  & 83.2 & 74.44\\
Cascade $\gamma_t^{cls}$ ~~~$\rightarrow$ $Q(p_t)$ & \textbf{74.47} & \textbf{84.33} & \textbf{75.93} \\
Cascade $Q(p_t)$ $\rightarrow$ $\gamma_t^{cls}$ & 73.0 & 82.75 & 74.55  \\
\bottomrule
\end{tabular}
}
\end{table}

Next we evaluate the pseudo-label elimination strategy of either a standard score threshold or quantile function, compared with the proposed cascade approach (\cref{eq:quantile_threshold}). Notably, setting the threshold or quantile too low would include more false positive predictions in training. Setting them too high would eliminate correct predictions since very few predictions would meet the required prediction score. This holds even when both threshold and quantile are dynamic (changing via predefined schedule). \cref{tab:abalation_threshold_cascade} shows that the cascade approach which first enforces a lenient threshold and then a quantile  yields the best results.

\begin{figure}[htbp]
    \centering
    \begin{subfigure}[b]{0.4\textwidth}
        \includegraphics[width=\textwidth]{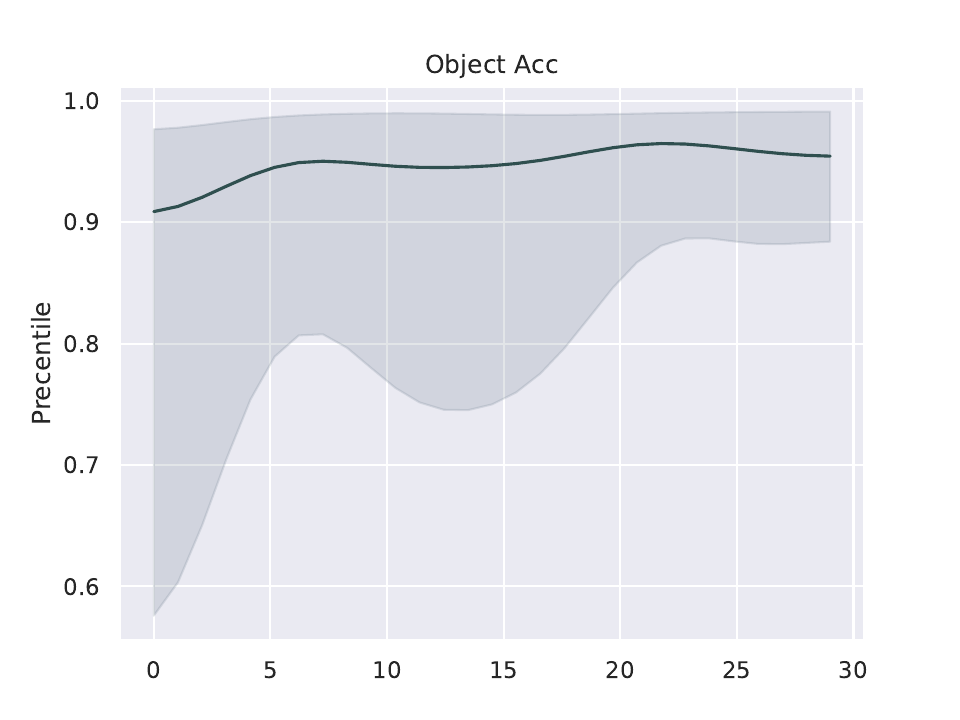}
        \caption{\textbf{Label accuracy over time} 
        The solid-black line represent the ``ideal'' class threshold $\gamma^{cls}$ that would eliminate all false-positive predictions and retain only true-positives. The regions highlighted in gray denote the standard-deviation of the ideal threshold.}
        \label{fig:class_acc}
    \end{subfigure}
    \hfill
    \begin{subfigure}[b]{0.4\textwidth}
        \includegraphics[width=\textwidth]{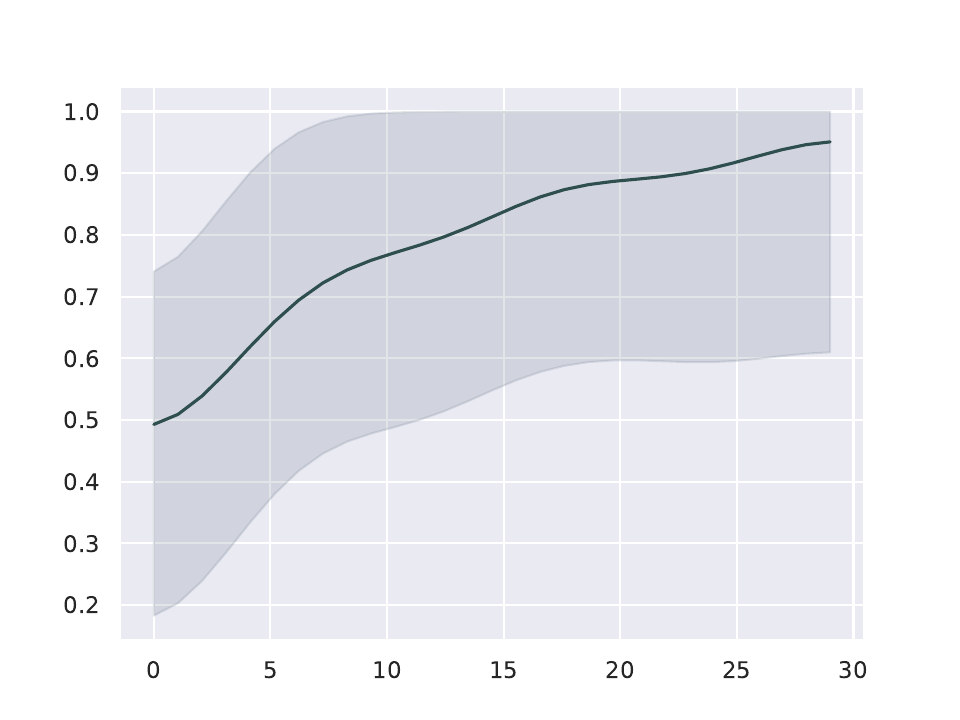}
        \caption{\textbf{Mask accuracy over time} 
        The solid-black line describes the mask threshold $\gamma^{mask}$ that for each instance, distinguish between pseudo-masks and background. The gray-highlighted regions denote the standard-deviation .}
        \label{fig:mask_acc}
    \end{subfigure}
    \caption{\textbf{Label and Mask Accuracy} of pseudo-labels. For both he $x$-axis measures training steps in multiples of $\times 1000$.}
    \label{fig:acc}
\end{figure}

In \cref{fig:acc} we illustrate that setting the threshold too high at any point during training would eliminate many true-positive predictions, solely due to the model predicting a low class label score. This is more prominent early in training, since the model gains confidence in its predictions as training progresses. We visualize the ``ideal'' threshold that would only retain true-positives and ensure that no false-positive pixels are passed through.

\begin{table}
\centering
\caption{
Ablation study of mask threshold $\gamma^{mask}$, class threshold $\gamma^{cls}$ and pseudo-box strategies, showing $\text{AP}$ for \textbf{1\% annotated data}. The Pseudo-box column corresponds to the standard pseudo-box approach which discards box predictions corresponding to pseudo-labels below the threshold $\gamma^{cls}$. The Mask-to-Box (M2B) and the combined Mask-to-Box with Multi-Label-Matching (M2B+MLM), introduced in this work, extract pseudo-boxes from pseudo-masks and instead filter individual pixels whose score fall below $\gamma^{mask}$. The table shows that M2B+MLM produced the best results.
}
\label{tab:ablation_box_threshold}

\resizebox{\linewidth}{!}{
\begin{tabular}{c|cccc}
\toprule
$\gamma^{cls}$ or $\gamma^{mask}$ & Pseudo-box & M2B  & M2B + MLM  \\
\toprule
0.5 & 71.98  & 72.71 & \textbf{73.00} \\
0.6 & 71.70  & 72.55 & \textbf{72.87} \\
0.7 & 71.57  & 72.28 & \textbf{72.86} \\
\bottomrule
\end{tabular}
}
\end{table}

Finally, we measure the impact of the different pseudo-box strategies. \cref{tab:ablation_box_threshold} shows the resulting AP for \methodname{} trained on $1\%$ of the labeled data and various values of threshold $\gamma^{mask}$. The standard approach filters the boxes by thresholding the class prediction score using $\gamma^{cls}$. We denote by M2B the effect of Mask-to-Box (extracting bounding boxes from the predicted instance masks), and denote by MLM the use of multiple boxes towards the self-supervised loss $\mathcal{L}_\mathrm{u}$ in \cref{eq:unsupervised_loss}.
The results demonstrate that employing a mask threshold of $\gamma^{mask}=0.5$ in conjunction with M2B + MLM achieves the best performance.

\section{Conclusion}
\label{sec:Conclusion}

In this work, we present \methodname{}, a novel framework that incorporates semi-supervised learning with learning through scene interaction in the context of a few-annotation data regime. \methodname{} is modular and can complement other segmentation models that emit intermediate instance embedding.
We demonstrate that \methodname{} improves $\text{AP}_{50}$ by over $+10$ compared to previous state of the art, after training end-to-end on just $0.5\%$ of the labeled data (with $99.5\%$ of the data treated as unlabeled). With just $1\%$ of the labeled data, \methodname{} achieves better performance than the baselines (DeDETR, SAM, RoboLLM) trained on $10\times$ the amount of labeled data. On OCID (RGB), \methodname{} sets a new state of the art, and is near state of the art when restricted to a fraction of the annotations.
For future work, we intend on leveraging ``before'' and ``after'' observations directly using robotic item grasping in real-world environments (rather than synthetically inserting instances into images), with the overarching goal of lifelong learning for robot perception.

A limitation of the proposed approach is that it underperforms when presented with objects that make few or no appearances in the truncated (labeled) training data. Access to a handful of annotated examples means that not all objects are encountered during training, resulting in some cases where two objects are segmented as one. In the context of robotic grasping, this may lead to a failed object-grasping attempt. However, since grasp failures also alter the scene, we believe that capturing snapshots of the scene before and after the attempted interaction would help improve the grasping precision in the long term.

{\small
\bibliographystyle{ieee_fullname}
\bibliography{egbib}
}

\clearpage
\setcounter{page}{1}

\noindent\section*{\centering Supplementary Materials for Robot Instance Segmentation with Few Annotations for Grasping}
\label{sec:supplementary}

\appendix
\section{Technical details}
\label{appdx:technical}

\noindent\textbf{Model} The \methodname{} framework begins with an image augmentation step that feeds into a feature extractor followed by an instance segmentation model, and ends at prediction heads for class, bounding box, mask and instance association. We use ResNet-50, ResNet-101~\cite{resnet} and Swin-L transformer ~\cite{liu2021Swin} as backbones throughout our experiments, followed by Deformable DETR ~\cite{zhu2020deformable} with $6$ encoders and decoders, width of $256$ and $300$ fixed instance queries, converging on an FPN-like dynamic mask head (as in SeqFormer \cite{seqformer}). In our evaluation, we measure the contribution of the proposed approach to Deformable DETR which serves baseline, and all feature extractors are pretrained on COCO instance segmentation, as is common in Instance segmentation pretraining ~\cite{Zoph2020RethinkingPA}. The proposed method incorporates a contrastive head (inspired by IDOL \cite{IDOL}) and introduces instance bank, self-supervision branch for non-labeled data, coupled prediction heads for stability (M2B) and label matching strategy during training (MLM). These, in aggregate, allow RISE to outperform both Deformable DETR and SAM, even when these are trained on $\times 10$ more data ($1\%$ vs $10\%$).

\noindent\textbf{Hyperparameters} Recall from \cref{sec:instance_segmentation} and \cref{sec:self_supervision} that the supervised loss $\mathcal{L}_\mathrm{s}$ and unsupervised loss $\mathcal{L}_\mathrm{u}$ (\cref{eq:supervised_loss}, \cref{eq:unsupervised_loss}, respectively) are a combination of the class loss $\mathcal{L}_{cls}$, bounding-box loss $\mathcal{L}_{box}$ weighted by $\lambda_1$, and the mask loss $\mathcal{L}_{mask}$ weighted by $\lambda_2$. We set the loss weights to be $\lambda_1 = 2.0$, $\lambda_2 = 1.0$. The total loss $\mathcal{L}_\mathrm{total}$ (in \cref{eq:total_loss}) combines the supervised loss $\mathcal{L}_\mathrm{s}$ or unsupervised loss $\mathcal{L}_\mathrm{u}$ (depending on availability of label $\mathbf{y}$), with association loss $\mathcal{L}_\mathrm{embed}$ weighted by $\lambda_3$. \cref{tab:abalation_IDLOSS} details an ablation of values of $\lambda_3$, showing that the $\mathcal{L}_\mathrm{embed}$ contributes to performance, with the best results attained for$\lambda_3=0.05$.

\begin{table}[h!]
\centering
\caption{Ablation of weight $\lambda_3$ applied to the sequence association loss $\mathcal{L}_\mathrm{embed}$ described in \cref{eq:loss_embed}. This evaluation uses $10\%$ of ARMBench labels ($90\%$ treated as unlabeled data) and the Swin-L as backbone. A value of $\lambda_3 = 0$ corresponds to a variant that ignores $\mathcal{L}_{embed}$. The best performance is obtained for $\lambda_3=[0.05,0.1]$.}
\setlength{\tabcolsep}{10pt}
\setlength{\abovecaptionskip}{-10pt}
\label{tab:abalation_IDLOSS}
\resizebox{0.99\linewidth}{!}
{
\begin{tabular}{l|ccc}
\toprule
$\lambda_3$ &$\text{AP}$ & $\text{AP}_{50}$  & $\text{AP}_{75}$\\
\toprule
0    & 74.7          & 84.9          & 75.9          \\ 
0.02 & 74.4          & 84.5          & 75.6          \\
0.05 & \textbf{74.9} & \textbf{85.2} & 76.0          \\
0.1  & 74.5          & 83.8          & \textbf{76.2} \\
0.5& 74.3            &84.2           & 75.3          \\
\bottomrule
\end{tabular}
}
\end{table}

\begin{figure}[h!]
  \centering
  \setlength{\belowcaptionskip}{0pt}
   \includegraphics[width=0.99\linewidth]{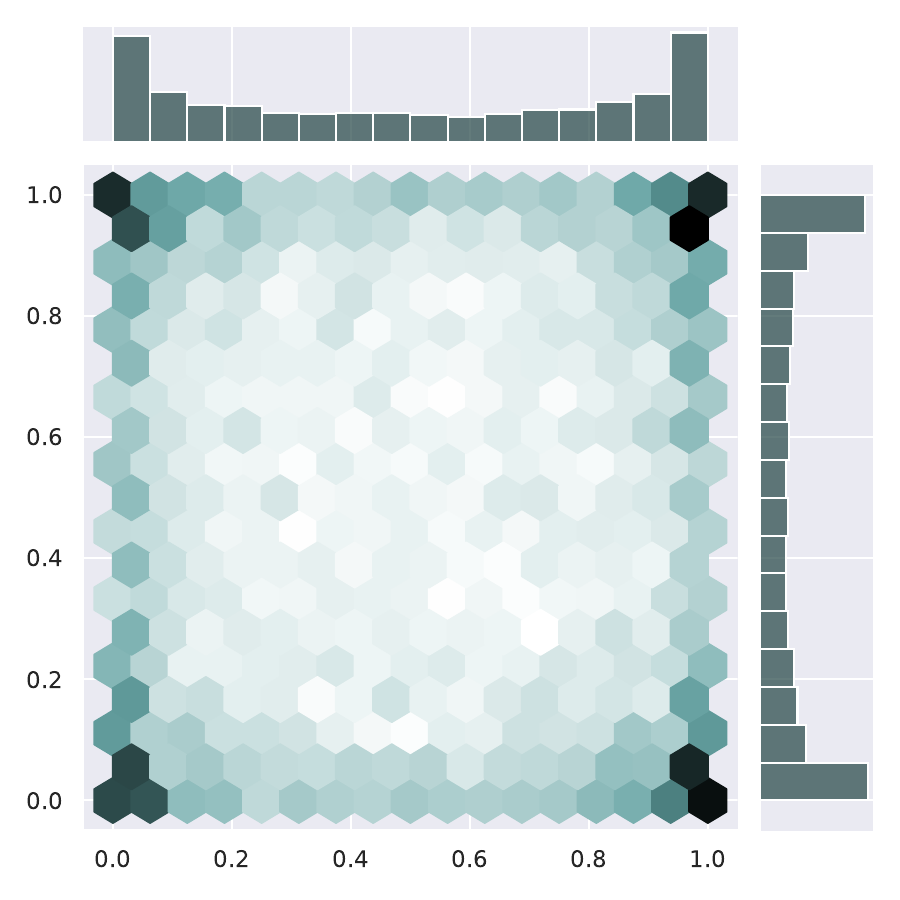}
   \caption{Two dimensional independent $Beta(\alpha=0.5, \beta=0.5)$ distribution representing the spread of instance-bank objects inserted into unlabeled images. The distribution favors placing inserted objects at the periphery of the image, since most images contain most of their information about their center (bright regions denote low probability).}
   \label{fig:beta}
\end{figure}

\noindent\textbf{Augmentation Strategy} The input images are downsampled and randomly cropped so that the longest side is at most $600$ pixels, and so that the shortest side is at least $480$ pixels. Recall from \cref{subsec:association_loss} that the instance bank contains object cutouts from the labeled portion of the dataset, inspired by \cite{dwibedi2017cut}. We randomly insert $K$ instances from the instance bank into the image to produce the ``before'' image $x_1$ and apply weak augmentations (e.g. slight rotation, translation, brightness etc.). However, we depart from previous approaches by having these $K$ instances distributed according to $Beta(\alpha=0.5, \beta=0.5)$, depicted in \cref{fig:beta}, making it less likely for synthetically placed instances to occlude actual objects in the image. In addition, we also ensure that the $K$ inserted instances don't overlap with one another beyond $85\%$ since they form ground truth labels during self-supervision. We then generate the ``after'' image $x_2$ by randomly adding more instances from the instance bank, or alternatively removing (or transforming) already inserted instances, followed by another round of weak augmentations. The before and after frames serve toward learning through interaction, and we facilitate self-supervised learning by strongly augmenting $x_1$ to yield $x_3$ and treat $x_w=x_1$ and $x_s=x_3$ as an input a pair of weakly- and strongly-augmented images. We employ this approach in our evaluation of both ARMBench \cite{Mitash2023ARMBenchAO} and OCID \cite{OCID} without the needing to tune its parameters specific datasets. 

\noindent\textbf{Training.}
We train our model for $12,000$ iterations, using AdamW~\cite{AdamW} optimizer with learning rate of $10^{-4}$, and weight decay of $10^{-4}$ and lr scheduler of StepLR that steps down an order of magnitude after $8,000$ iterations.

\section{Prediction Matching}
The model predicts up to $300$ instance labels, boxes and masks which are often far beyond the actual instance count in a given image. In order to compute the loss between valid predictions and ground-truth annotations, we compute the bipartite cost matrix which measures the IoU of each prediction against each ground-truth annotation (either based on box IoU or using the Mask-to-Box method detailed in \cref{sec:self_supervision}). We then find the fitting assignment for each ground-truth annotation by solving an Optimal Transport (OT) Problem \cite{Ge2021OTAOT}. A similar approach described in \cref{par:Association_Loss} serves toward computing $\mathcal{L}_{\mathrm{embed}}$ which requires positive and negative views of an instance. We introduce a method inspired by IDOL \cite{IDOL}, where the top-$10$ prediction matches of each ground-truth annotation are treated as positive views and the rest are considered negative views. The impact of matching is evident in the ablation study in $\cref{tab:abalation_components}$ where we use either OT or a more standard approach of using the top $0.7$ IoU as positive and bottom $0.3$ IoU as negative.

This flow is similarly applied during the self-supervision phase, with the distinction of using pseudo- labels, boxes and masks instead of manually annotated ground truth. Here we also employ Multi-Label Matching (MLM, \cref{sec:self_supervision}) to allow the model to learn from multiple pseudo-labels predicted from the weak augmentation $x_w$. The impact of MLM is demonstrated in $\cref{tab:abalation_components}$ and inspired by \cite{Chen2022LabelMatching}, where it further contributes to the framework's performance.

\section{Thresholds}
\label{appx:threshold_info}
We use time-dependent thresholds \cite{zhang2021flexmatch,kim2021selfmatch}, whereby an initial threshold value  increases every $1000$ training steps. 
The class and mask thresholds start at $\gamma_t^{\text{cls}}=\gamma_t^{\text{mask}}=0.85$ and peak at $0.98$. For the Cascade approach (\cref{sec:self_supervision}) which combines a lenient threshold followed by a quantile $Q_t$ described in \cref{eq:quantile_threshold}. We set the initial class and mask thresholds to be $\gamma_t^{\text{cls}}=\gamma_t^{\text{mask}}=0.5$ and peak at $0.85$. The quantile $Q_t$ follows the schedule $p_t = a_0 \cdot (1 - \nicefrac{t}{T})$ where $t$ is the training step, $T$ denotes the total number of training steps, and $a_0=0.995$ is the quantile base value. Upon ranking the model's predicted instances by their class score, only the top $p_t$ are retained, and the rest are discarded.

\section{Foundation Model Comparison}
\label{paragraph:foundation_models}
\begin{figure}
  \setlength{\belowcaptionskip}{0pt}
  \centering
   \includegraphics[width=.8\linewidth]{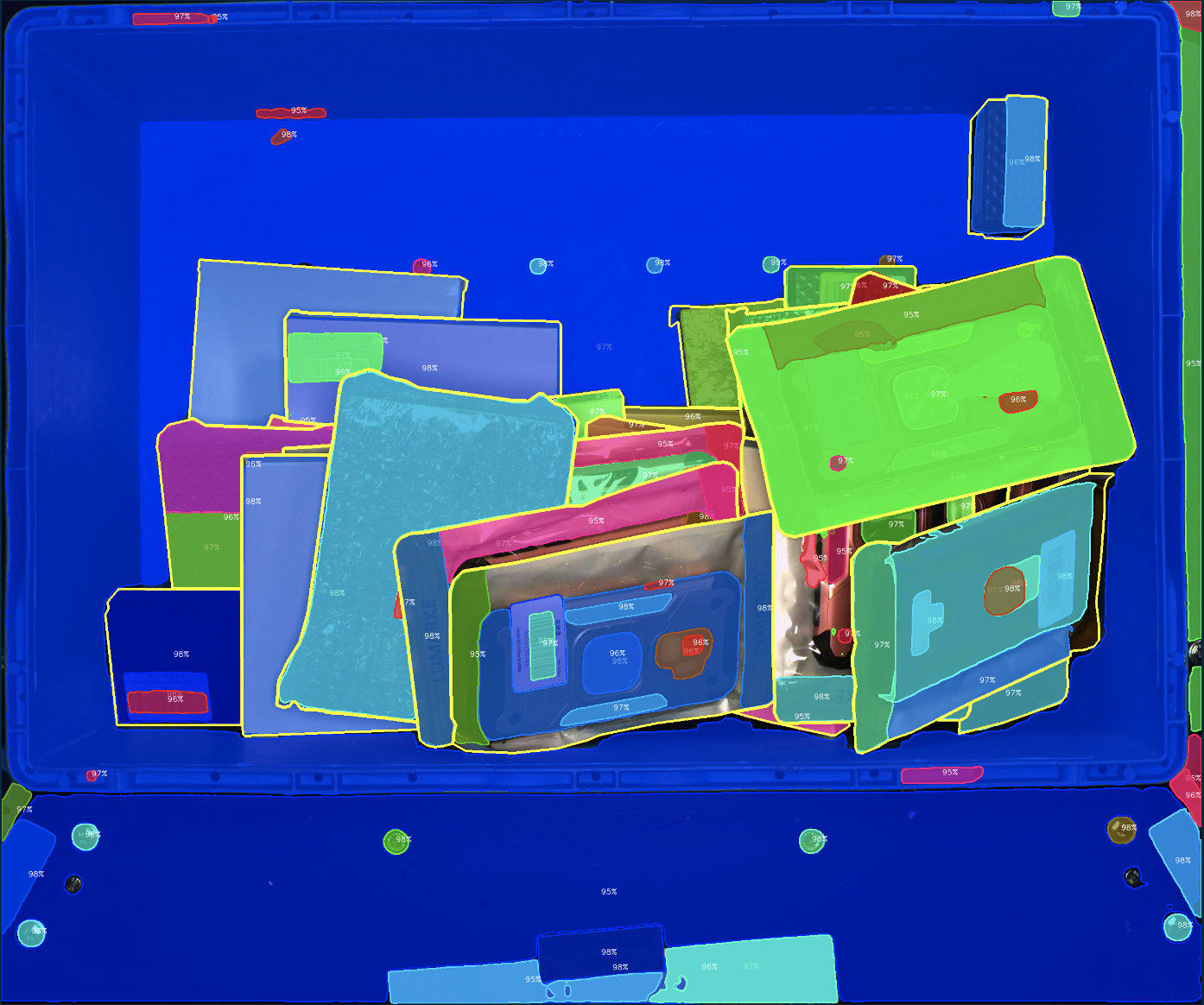}
   \caption{Fine-tuned SAM ($1\%$ ARMBench data) showing many fragmented masks and false positive artifacts.}
   \label{fig:sam}
\end{figure}

As an additional baseline we compare \methodname{} with the ``Segment Anything'' (SAM) foundation model \cite{kirillov2023segany}, fine-tuned on a subset of the ARMBench dataset.
In \cref{tab:classic} we demonstrate that despite SAM's unrivalled ability to \textit{segment anything}, it is prone to over-segment and produce mask artifacts, even after fine-tuning on a small portion of domain-specific images. \cref{fig:sam} shows an example where SAM, fine-tuned on $1\%$ of the data still struggles with accurately discerning objects, resulting in fragmented and incomplete object masks and mask predictions that target less significant elements of the image (such as packaging features, rivets and shadows). The numerous false positive predictions impact the overall performance.

\begin{figure*}

    \centering
    \begin{subfigure}{0.19\linewidth}
    \includegraphics[width=\textwidth]{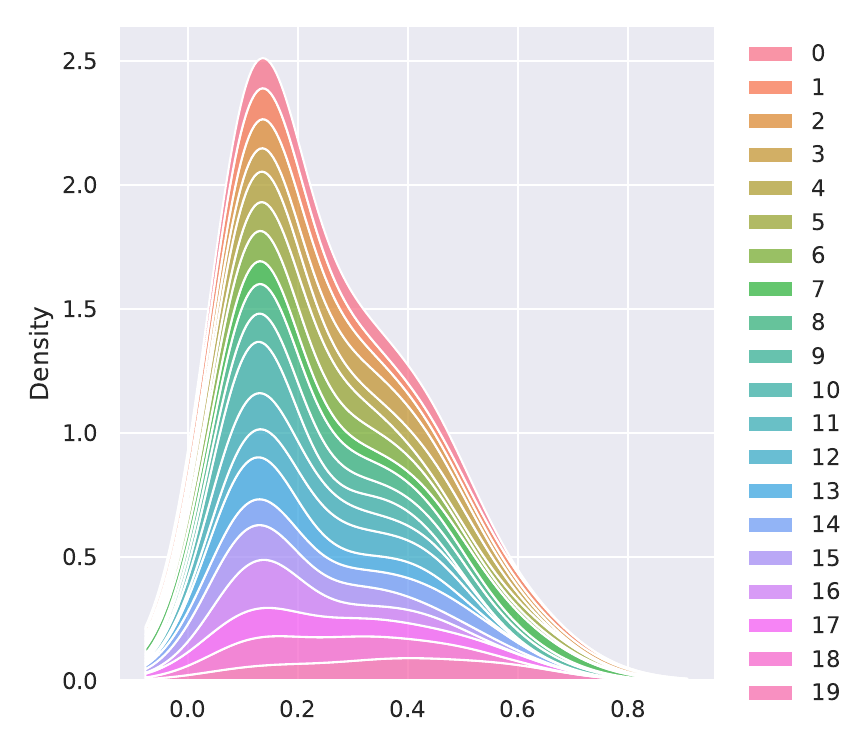} 
        \caption{$a_0$ = 0.7}
    \end{subfigure}
    \begin{subfigure}{0.19\linewidth}
 	\includegraphics[width=\textwidth]{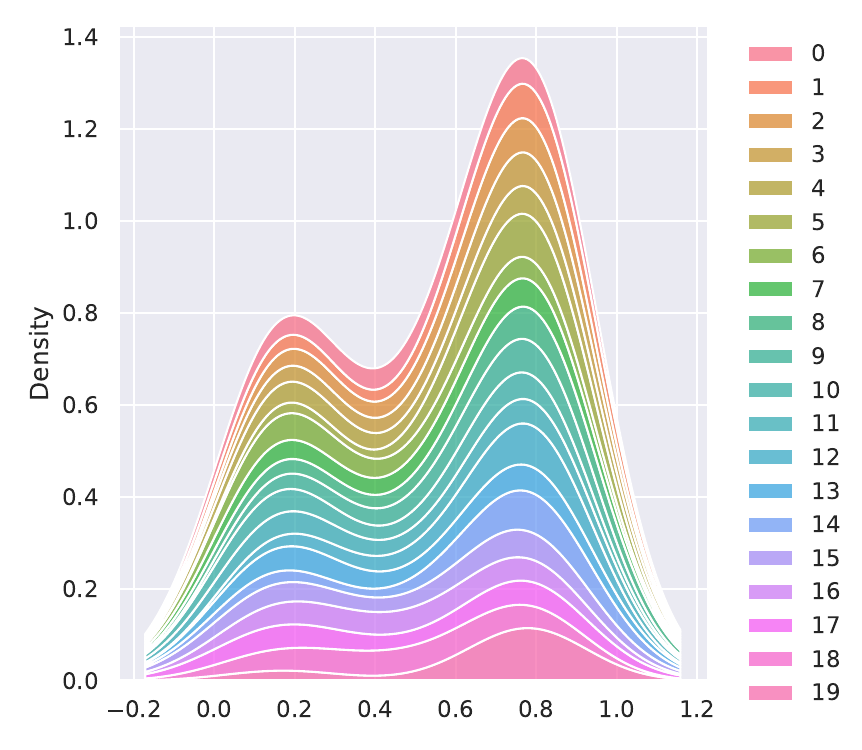}
        \caption{$a_0$ = 0.9}
    \end{subfigure}
    \begin{subfigure}{0.19\linewidth}
 	\includegraphics[width=\textwidth]{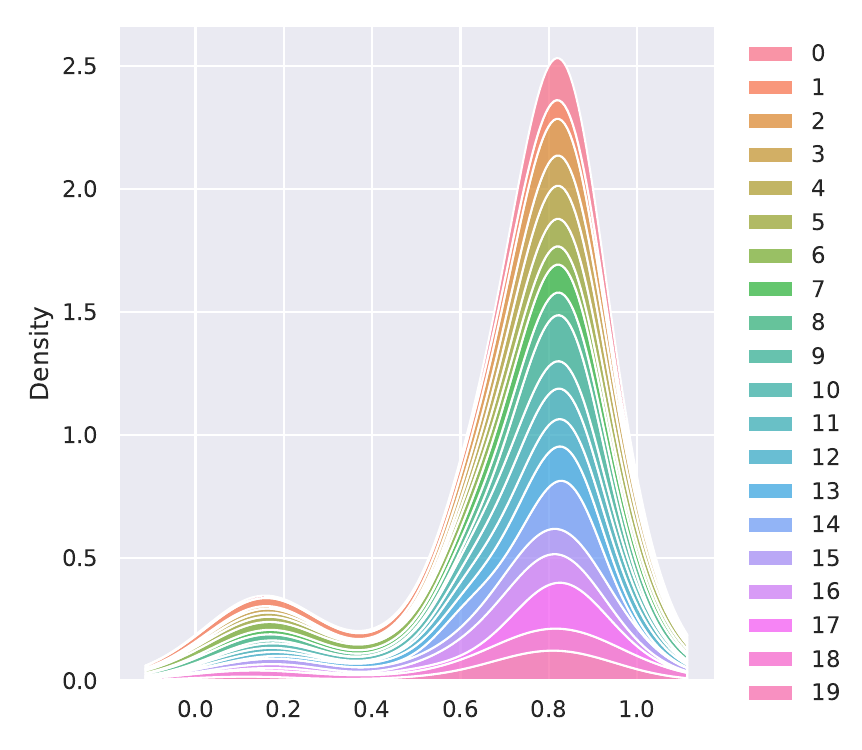}
        \caption{$a_0$ = 0.95}
    \end{subfigure}
    \begin{subfigure}{0.19\linewidth}
 	\includegraphics[width=\textwidth]{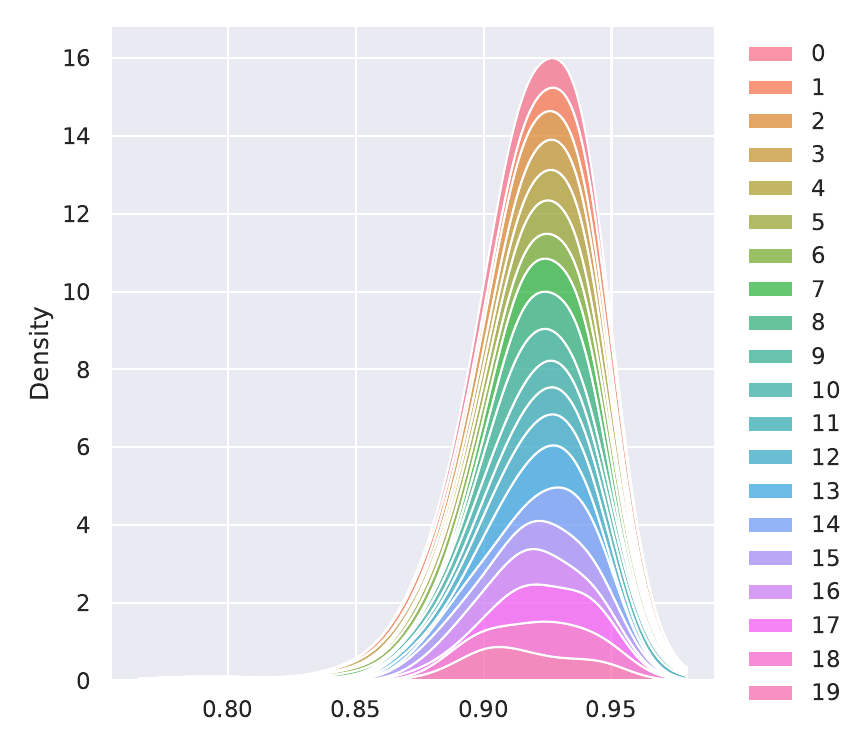}
        \caption{$a_0$ = 0.99}
    \end{subfigure}
    \begin{subfigure}{0.19\linewidth}
 	\includegraphics[width=\textwidth]{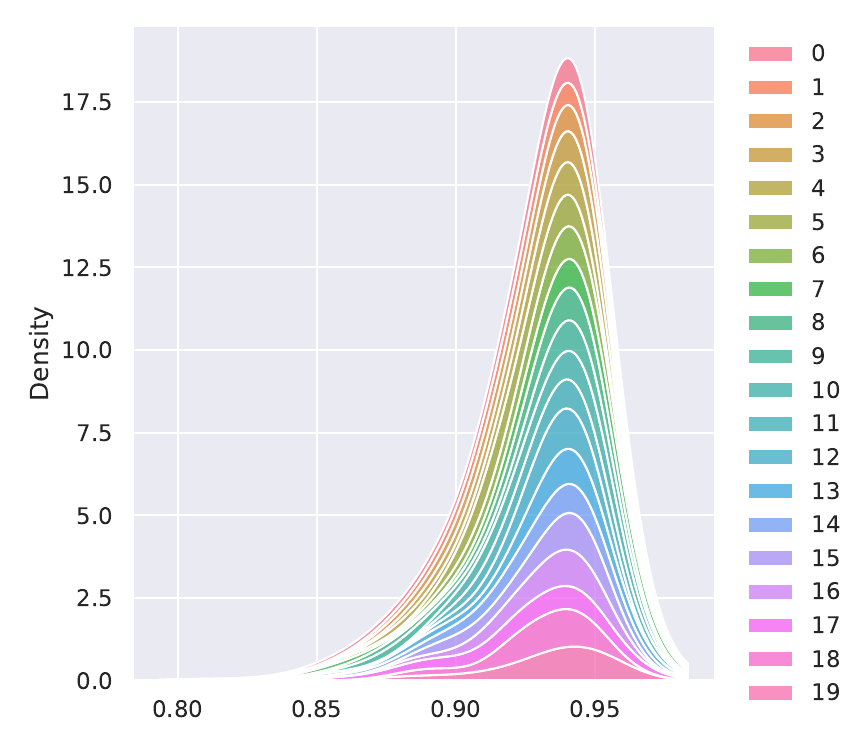}
        \caption{$a_0$ = 0.995}
    \end{subfigure}
    \setlength{\belowcaptionskip}{-2pt}
	\caption{Confidence density over time using different quantile values for the cascade threshold (\cref{eq:quantile_threshold}). The $x$-axis represent the score of all samples, the $y$-axis the valid instance-count (instance density), and the color band correspond to training iterations in increments of $1000$. The cascade threshold applies both a time-dependent threshold (which tightens over time) and a time-dependent quantile $Q_t$ (which loosens over time). The base quantile value $a_0$ is detailed for each subfigure, showing that the best initial value for the quantile is $0.995$.}
	\label{fig:DPA_density}
\end{figure*}

\section{Study of Cascade Threshold}
\label{sec:dynamic_threshold}
We study the behavior of pseudo-label and pseudo-mask Cascade filter strategy (\cref{eq:quantile_threshold}). We evaluate the per-instance prediction score of the model using different base values for the quantile $Q_t$ of the Cascade threshold. In \cref{fig:DPA_density}, each color band represent $1000$ iterations. The figure shows that setting the  base value of the quantile too low would allow in more false-negatives as pseudo- labels and masks. Alternatively, setting it too high would discard valuable predictions as they don't meet the ranking requirement of the qunatile. Following this evaluation we set the quantile base value to $a_0 = 0.995$, which leads to the most balanced behavior of discarding false-positive predictions while allowing through true-positives (even when their score would be considered too low by a standard scheduled threshold).

\section{Failure Cases}
\begin{figure*}[hb!]
\includegraphics[width=0.9\textwidth]{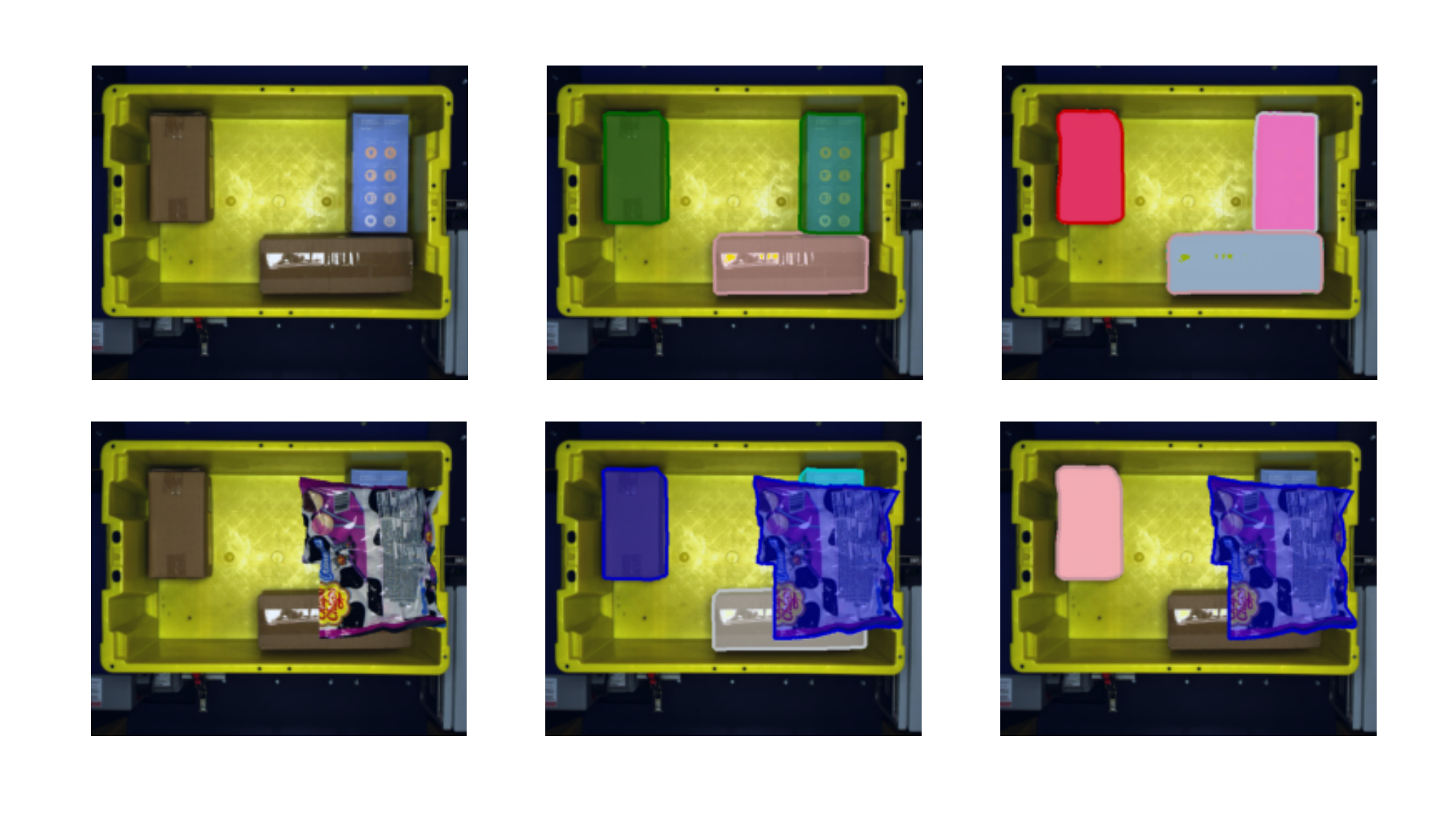}
\centering
\caption{ \textbf{Failure cases.} Mask prediction of a model trained on $1\%$ of annotated data and $99\%$ unlabeled (ResNet-50 backbone). The top row shows that the model accurately predicts the three objects in the tote. The bottom row includes an additional item inserted from the instance-bank, which partially overlaps several objects. Although the model correctly segments the inserted object, it completely misses one occluded object.
}
 \label{fig:failed}
\end{figure*}

In both the supervised and self-supervised stages, we randomly draw instance-bank objects and distribute them in the image according to a 2d $Beta(\alpha,\beta)$ distribution (\cref{eq:beta}), and prevent object overlap that exceeds $85\%$ by resampling from the distribution in case of such overlap.
In the supervised phase, we also ensure that inserted objects do not overlap existing (ground-truth) objects by more than $85\%$, whereas in the self-supervised phase, the $Beta(\alpha,\beta)$ distribution (\cref{eq:beta}) helps reduce the likelihood of inserted memory-bank objects overlapping actual objects in the image (since no ground-truth is available).
Despite these precautions, failure cases still occur, particularly at very low annotation rates. 
Since our method incorporate noisy pseudo--labels in low annotated data regime, we will follow improvements in noisy spatial labels \cite{grad2024benchmarking} for combating with noise and improve pseudo--labels.
\cref{fig:failed} shows how a model trained on $1\%$ of the labeled data ($99\%$ treated as unlabeled) accurately predicts the masks of all objects in the ``before'' image $x_1$ (and ignores the background). However, in the ``after'' image $x_2$ (post-interaction), which contains an additionally inserted object (bottom row), the model fails to produce masks for the occluded cardboard box.\\

\noindent \textbf{Acknowledgment:}
Bin illustrations in figure \cref{fig:rise_architecture} are attributed and modified from  \cite{ycb}.

\end{document}